\documentclass[11pt,a4paper]{article}
\usepackage[dvipsnames]{xcolor}
\usepackage[hyperref]{emnlp2021}
\usepackage[export]{adjustbox}
\usepackage{appendix}
\usepackage{booktabs}
\usepackage{rotating}
\usepackage{graphicx}
\usepackage{latexsym}
\usepackage{linguex}
\usepackage{multirow}
\usepackage{tabularx}
\usepackage{times}
\usepackage{url}

\usepackage{cleveref}
\usepackage[textsize=huge,disable]{todonotes}

\usepackage{microtype}

\newcommand{\note}[4][]{\todo[author=#2,color=#3,size=\scriptsize,fancyline,caption={},#1]{#4}} 

\newcommand{\adrian}[2][]{\note[#1]{adrian}{red}{#2}}

\newcommand{\NoteAB}[2][]{\adrian[inline,#1]{#2}\noindent}

\newcommand{\removed}[1]{}
\newcommand{\REMOVED}[1]{}

\newcommand{\adj}[0]{\texttt{ADJ}}
\newcommand{\adv}{\texttt{ADV}}

\newcommand{\noun}{\texttt{NOUN}}
\newcommand{\propn}{\texttt{PROPN}}
\newcommand{\posverb}{\texttt{VERB}}
\newcommand{\adp}{\texttt{ADP}}
\newcommand{\aux}{\texttt{AUX}}
\newcommand{\cconj}{\texttt{CCONJ}}
\newcommand{\posdet}{\texttt{DET}}
\newcommand{\num}{\texttt{NUM}}
\newcommand{\pospart}{\texttt{PART}}
\newcommand{\pron}{\texttt{PRON}}
\newcommand{\sconj}{\texttt{SCONJ}}
\newcommand{\punct}{\texttt{PUNCT}}
\newcommand{\sym}{\texttt{SYM}}

\definecolor{PredicateColor}{RGB}{255,102,102}

\newcommand{\predicate}[1]{$[_{P \,}  $\textbf{\underline{\color{PredicateColor} #1}}$]$}
\newcommand{\argone}[1]{$[_{A1 \,}  $\emph{\color{Blue} #1}$]$}
\newcommand{\argtwo}[1]{$[_{A2 \,}  $\emph{\color{YellowOrange} #1}$]$}

\newcommand{\dlurl}[0]{\url{https://zenodo.org/record/5495668}}

\definecolor{shadecolor}{RGB}{240,255,240}
\definecolor{bluebgcolor}{RGB}{210,236,247}

\newcommand{\shady}[1]{\par\noindent\colorbox{bluebgcolor}{\parbox{\dimexpr\textwidth-2\fboxsep\relax}{#1}}}

\title{Cross-Register Projection for Headline Part of Speech Tagging}

\author{
  Adrian Benton, Hanyang Li, Igor Malioutov  \\
  Bloomberg \\
  731 Lexington Ave \\
  New York, NY 10022 USA \\
  {\tt \{abenton10,hli762,imalioutov\}@bloomberg.net}}

\date{}

\begin{document}
\maketitle
\begin{abstract}
Part of speech (POS) tagging is a familiar NLP task.  State of the art taggers routinely achieve token-level accuracies of over 97\% on news body text, evidence that the problem is well-understood.  However, the register of English news headlines, ``headlinese'', is very different from the register of long-form text, causing POS tagging models to underperform on headlines.  In this work, we automatically annotate news headlines with POS tags by projecting predicted tags from corresponding sentences in news bodies.  We train a multi-domain POS tagger on both long-form and headline text and show that joint training on both registers improves over training on just one or na\"{i}vely concatenating training sets.  We evaluate on a newly-annotated corpus of over 5,248 English news headlines from the Google sentence compression corpus, and show that our model yields a 23\% relative error reduction per token and 19\% per headline.  In addition, we demonstrate that better headline POS tags can improve the performance of a syntax-based open information extraction system.  We make \texttt{POSH}, the POS-tagged Headline corpus, available to encourage research in improved NLP models for news headlines.
\end{abstract}

\section{Introduction}
\label{sec:introduction}

News headlines were identified as a unique register of written language at least as far back as \newcite{straumann1935newspaper}, and the term \emph{headlinese} was coined specifically to refer to the unnatural register used in headlines.  Hallmarks of headlinese include frequent omission of articles and auxiliary verbs, stand-alone nominals and adverbials, and infinitival forms of the main verb when referring to the future tense \cite{maardh1980headlinese}.  In spite of the clear differences in syntax between news headlines and bodies, the NLP community has invested little effort in building headline-specific models for predicting syntactic annotations such as part of speech (POS) tags and dependency parse trees. 
This limits the use of syntax-based models on headlines.  For example, the Open Domain Information Extraction (Open IE) system, PredPatt \cite{white2016universal}, cannot be expected to perform well on headlines, if existing models cannot accurately tag headlines with POS tags and dependency relations. More broadly, headline processing provides an important signal for downstream applications such as summarization ~\cite{bambrick-etal-2020-nstm}, sentiment/stance classification~\cite{strapparava2007semeval,ferreira2016emergent}, semantic clustering \cite{wities2017consolidated,laban2021news}, and information retrieval~\cite{lee2010mining}, among other applications. 



\begin{figure*}[!ht]
{
\renewcommand{\arraystretch}{1.5}%
\fontfamily{phv}\selectfont%
\large%
\shady{
\begin{tabularx}{\linewidth}{l}
  Japanese Firms Push Posh Car Showrooms\\
  Twinkies are Back\, but Smaller\\
  World Briefing | Africa: Angola: New Bid To End War\\
  Ebola Fear Stalks Home Hunt for Quarantined Now Released\\
  Madonna Addicted to Sweat Dance Plugs Toronto Condos: Mortgages\\
\end{tabularx}
}}
\caption{Examples of headlines exhibiting register-specific phenomena: article/auxiliary omission, multiple ``decks'' or independent segments, and garden paths following from the need to fit a large amount of information in a small space.}
\end{figure*}

\removed{
\begin{figure*}[ht]
    {\fontfamily{qcr}\selectfont
\shady{
\textbf{
\renewcommand*{\familydefault}{\ttdefault}
\frenchspacing
\raggedright
\begin{itemize} \itemsep 0em
  \item[] Japanese Firms Push Posh Car Showrooms
  \item[] Twinkies are Back, but Smaller
  \item[] World Briefing | Africa : Angola : New Bid To End War
  \item[] Ebola Fear Stalks Home Hunt for Quarantined Now Released
  \item[] Madonna Addicted to Sweat Dance Plugs Toronto Condos: Mortgages
\end{itemize}
}
}
}
\caption{Examples of headlines exhibiting register-specific phenomena: article/auxiliary omission, multiple ``decks'' or independent segments, and garden paths following from the need to fit a large amount of information in a small space.}
\end{figure*}
}

In this work we take a first step toward developing strong NLP models for headlines by focusing on improving POS taggers.  We propose a projection technique (inspired by work on cross-lingual projection ~\cite{yarowsky-etal-2001-inducing}) to generate silver training data for each headline based on POS tags from the lead sentence of the news body. To facilitate evaluation, we crowdsource and manually adjudicate gold POS tag annotations for 5,248 headlines from the Google sentence compression corpus \cite{filippova2013overcoming}.  To the best of our knowledge, this is the first headline dataset with gold-annotated POS tags.

We evaluate a range of neural POS taggers trained on long-form text, silver-labeled headlines, and the concatenation of these datasets, and find that multi-domain models which train a separate decoder layer per domain outperform models trained on data from either headlines or long-form text.  
We finally show that more accurate POS taggers directly lead to more precise Open IE extractions, yielding an extracted tuple precision of 56.9\% vs. 29.8\% when using a baseline POS tagger trained on the English Web Treebank alone. We release our gold-annotated data as \texttt{POSH}, the POS-tagged Headline corpus, 
at \dlurl.

\paragraph{Contributions}

\begin{enumerate} \itemsep 0.2em
 \item A gold-annotated evaluation set of over 5,000 English headlines to promote stronger NLP tools for news headlines. 
 \item An error analysis, confirming that many of the errors made by taggers trained on body text are due to headline-specific phenomena, such as article and copula omission. This demonstrates that existing POS taggers are not effective at headline POS tagging.
 \item By training models on headlines tagged using a simple projection of POS tags from the corresponding news body's lead sentence, we can outperform a tagger trained on gold-labeled long-form text data.  Multi-domain taggers that are trained on both gold-annotated long-form text and silver-labeled headlines outperform taggers trained on either, reducing the relative token error rate by 23\%. 
 \item Finally, we demonstrate that more accurate headline POS tags translate to more precise tuple extractions in a state-of-the-art multilingual Open IE system.
\end{enumerate}

The proposed projection technique and models can also be applied to other sequence tagging tasks such as chunking or named entity recognition. These are also applicable in other domains where one has access to long-form sentence and a parallel reduced sentence (e.g., simplified English \cite{coster2011simple}  or writing from English as a second language students \cite{dahlmeier2013building}).


\section{Data}
\label{sec:data}

We rely on two main sources of data for training and evaluating headline POS taggers: version 2.6 of the Universal Dependencies English Web Treebank 
(EWT; \citealt{silveira2014gold})\footnote{\url{https://github.com/UniversalDependencies/UD_English-EWT/tree/r2.6}}
and headlines from the Google sentence compression corpus (GSC).

In this work, we only consider the UD POS tag set.  UD has become a dominant tag set for tagging and parsing, with many treebanks available across many languages \cite{zeman2017conll}. In addition, we chose to annotate headlines with UD tags as the coarser granularity made it easier for non-experts to label.  
We leave experiments with finer representation granularity as future work, but do not expect the choice of representation to alter the underlying conclusions.

\NoteAB{Frame need for strong POS tagger by PredPatt earlier?  Citation for why UD vs. PTB?}

\subsection{Headline Evaluation Set}
\label{subsec:headline_eval_set}

We construct the GSC headline evaluation set (GSCh) by sampling uniformly at random from headlines in the GSC where the headline was a (possibly non-contiguous) subsequence of the associated lead sentence. 
Headlines were first tokenized using the Stanford CoreNLP PTBTokenizer \cite{manning2014stanford} with default settings, and were annotated by a pool of six annotators who were all proficient in English.  Each headline was annotated independently by three annotators from this pool.  Annotators were warm-started with POS tags generated by an EWT-trained BiLSTM model and were trained  to follow the UD 2.0 POS tagging guidelines.  When in doubt, annotators referred to similar examples in the UD 2.6 EWT.  Full annotation instructions and a sample of the interface are given in Appendix \ref{app:annotation_guidelines}.

We gave the annotators feedback by monitoring their performance on a set of 250 unique test examples randomly inserted as tasks (annotated by the authors).  We identified common mistakes on these questions and provided feedback to annotators throughout the annotation process, after every 500-1000 examples.

Annotators achieved unanimous agreement on 70.7\% of headlines, and the majority vote label sequence agreed with our test examples 88\% of the time.  We annotated an additional 200 samples from the unanimously agreed set, and found that 97\% of labeled examples completely agreed with our manual annotation.  We thus focused on those headlines without unanimous agreement, and manually reviewed the majority vote tag sequence for all headlines where there was no unanimous agreement (1534 out of 5248 examples).  We corrected at least one POS tag in 24.97\% of these headlines.  Common mistakes include labeling a nominalized verb as \posverb, labeling a \noun{} as \adj{} when it is a nominal modifier, confusing \adp{} and \sconj, and labeling ``to'' as \adp{} instead of \pospart.

\NoteAB{Include a data statement in appendix?  May not be necessary since these headlines are already released by GSC.}

\subsection{Headline vs. Body POS Tag Distribution}
\label{subsec:data_stats}

\begin{figure}
    \centering
    \includegraphics[width=1.0\linewidth,page=1,trim={1.05cm 0.72cm 0.72cm 0.7cm},clip]
    {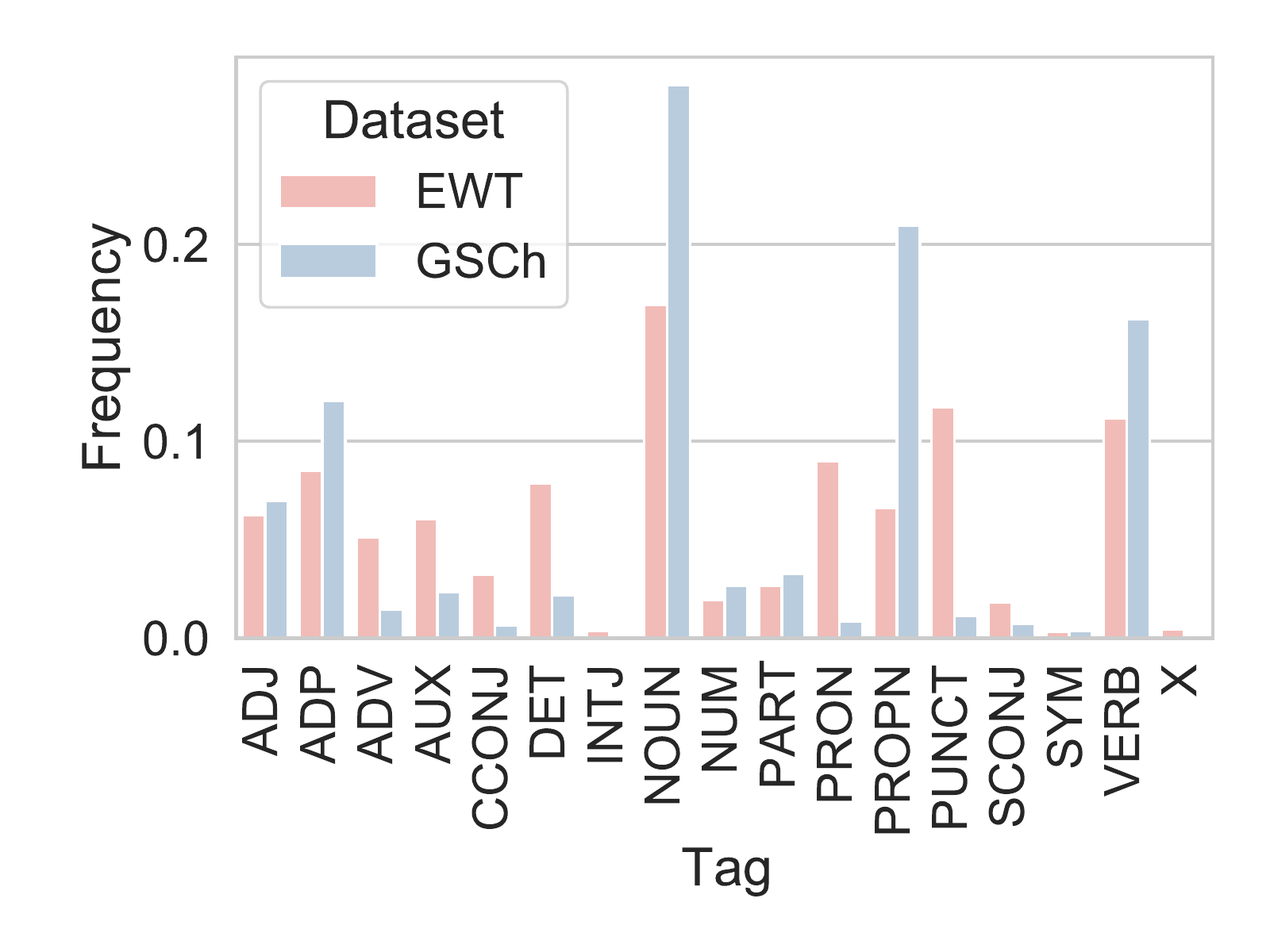}
    \caption{Unigram POS tag distribution for the EWT and GSCh datasets.}
    \label{fig:ewt_gsch_tag_unigram_stats}
\end{figure}

The unigram tag distribution for EWT and GSCh exhibit clear, expected differences (\Cref{fig:ewt_gsch_tag_unigram_stats}).  For instance, the lack of \posdet{} is due to article dropping and the lack of \adv{} and increase in frequency of other open class tags follows from the function that news headlines serve: maximizing the relevance of the article to a reader under strict space constraints \cite{dor2003newspaper}. Also, \propn{} is more frequent in headlines, while \pron{} is almost non-existent.  This naturally follows from the fact that headlines do not offer context from which an antecedent for the \pron{} can be found.  Type-token ratio is higher for GSCh than EWT (0.273 vs. 0.076), and GSCh headlines tend to be shorter than EWT sentences (7 vs. 15 mean token length).

\NoteAB{Leslie suggested looking at tag transition probabilities.  How else can we understand how headlines differ from body text?  Words in less frequent nominal sense (e.g., "estimates", "crash") will show up in headlines more often.}

\subsection{Auxiliary Datasets}
\label{subsec:aux_data}

We use the EWT as our baseline training set as the corpus contains over 250,000 annotated words of web text drawn from a wide variety of sources.  In addition, we use the Revised English News Text Treebank~\cite{ptb}, converted to Universal Dependencies using the Stanford CoreNLP Toolkit~\cite{manning+14}, GUM \cite{zeldes2017gum}, and English portions of the LinES \cite{ahrenberg2015converting} as additional training data for a subset of taggers.

In addition to the GSCh, we collected POS tag annotations for 500 additional headlines: 271 sampled uniformly at random from the GSC and 229 from The New York Times Annotated Corpus (NYT; \citealt{sandhaus2008new}). NYT headlines were restricted to those with 4-12 words (10$^{th}$-90$^{th}$ percentile of length distribution).  No subsequence constraint was imposed on any of these headlines, as we used this set to evaluate how well models performed on general headlines.

\section{Methods}
\label{sec:methods}

Here we describe the architecture of the POS taggers and our approach for learning a headline POS tagger without direct supervision.

\subsection{Models}
\label{subsec:models}

\begin{figure*}[t]
    \centering
    \includegraphics[width=0.9\linewidth,page=3,trim={0 1cm 0cm 0cm},clip]{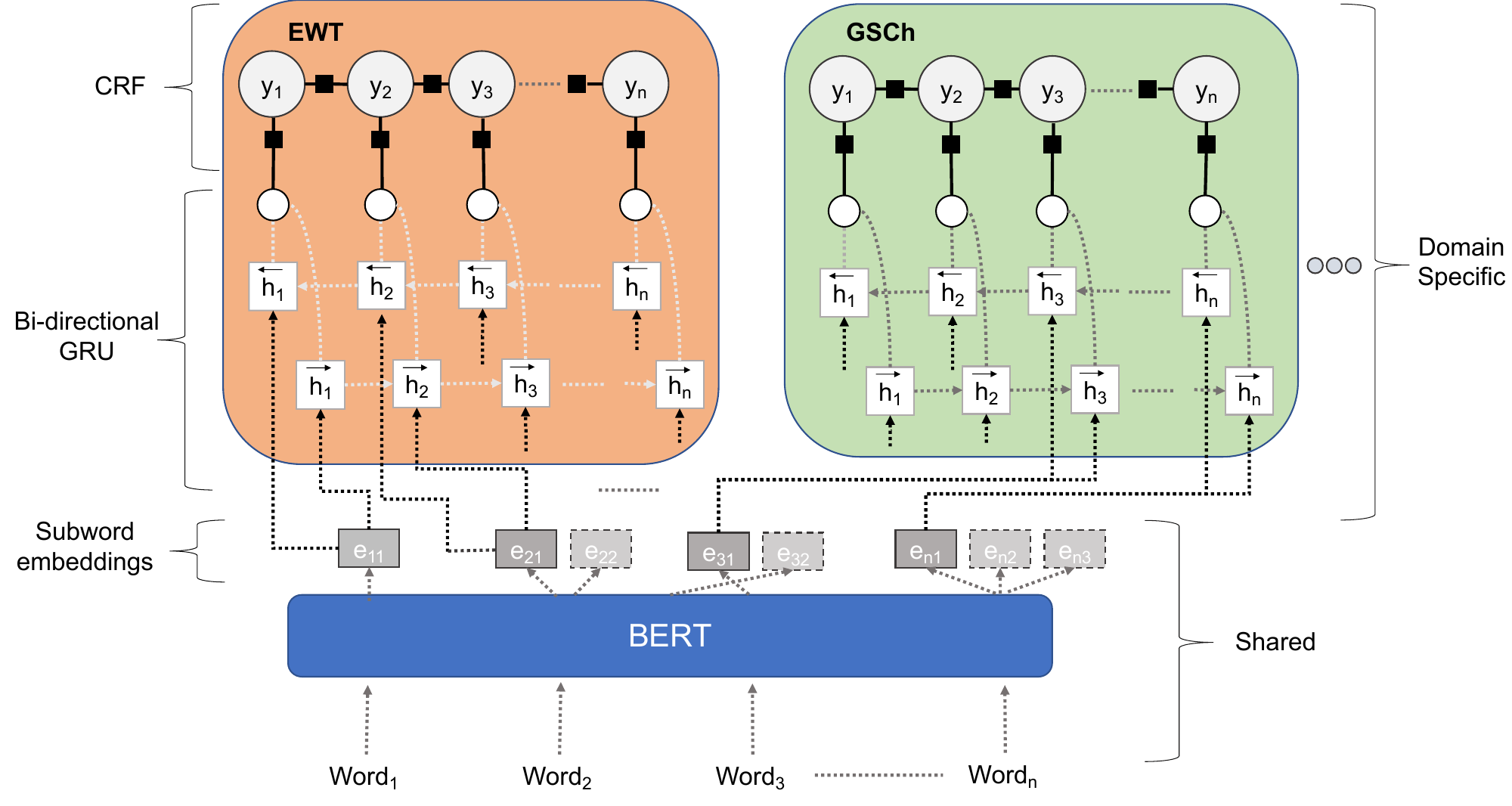}
    \caption{Schematic of the architecture for a multi-domain BERT POS tagger.  Word embeddings are given as the first subword embedding from an uncased BERT base language model.  
    The non-contextual variant is identical except it uses GloVe embeddings as input, a 2-layer RNN shared across domains, and only the tag projection and neural CRF layers are domain-specific.  Choice of decoder is governed by the corpus the example comes from.
    }
    \label{fig:tagger_architecture}
\end{figure*}

In all experiments, we use a bidirectional recurrent neural network with gated recurrent units, followed by a linear-chain conditional random field layer (BiGRU-CRF).  We consider two flavors of tagger: \emph{non-contextual} and \emph{contextual}.   The contextual taggers use the BERT \cite{devlin2019bert} base uncased model pretrained on Wikipedia and the Google books corpus as a token encoder,\footnote{We use the BERT implementation and model exposed by \url{https://huggingface.co/}.} and continue to finetune model weights during tagger training.   We represent each word by the embedding of its initial subword as done in \newcite{devlin2019bert}. The non-contextual taggers use 50-dimensional GloVe word embeddings \cite{pennington2014glove} concatenated with a 50-dimensional cased character embedding generated by a single-layer BiGRU.\footnote{\url{http://nlp.stanford.edu/data/glove.6B.zip}}  Non-contextual taggers use two-layer BiGRUs whereas contextual taggers only contain single-layer BiGRUs.  This is the same architecture used by \newcite{lample2016neural} for named entity recognition except with a GRU instead of a LSTM as the recurrent unit.\footnote{These settings of number of layers, embedding, and hidden layer dimensionality were found to perform well in preliminary experiments.  In addition, we saw little difference in performance between the two types of recurrent cells, and GRUs contain fewer parameters.}  Viterbi decoding is used to generate predictions for all taggers.

Even though large, pretrained language models have become a standard solution for addressing various NLP tasks, we also evaluate RNN taggers as they contain many fewer parameters and may suffice for POS tagging.  This is also a new task, and the data come from a domain which BERT was not explicitly trained on (news headlines). Therefore, it is not immediately clear that using a BERT POS tagger will outperform an RNN tagger.

\subsubsection{Multi-Domain Taggers}
\label{subsubsec:multidomain_tagger}

In our experiments, we also train taggers on data from multiple domains.
For the purpose of modeling, we designate each dataset as belonging to a separate \emph{domain}, even though the critical distinction between datasets may be due to a variation in register, not the subject matter.
In addition to training mixed-domain taggers on a simple concatenation of datasets, we consider tagger variants with domain-specific decoder layers.

Non-contextual, multi-domain taggers have domain-specific weights for the tag projection layer and CRF, but share the bidirectional GRU encoder across domains.  Contextual taggers share the BERT encoder across domains but learn domain-specific weights for the BiGRU-CRF layers.  This multi-domain architecture is a simpler version of that proposed by \newcite{peng2017multi}, since all models are trained to perform one task, POS tagging (\Cref{fig:tagger_architecture}).

\subsection{Projecting POS Tags}
\label{subsec:projecting_pos_tags}

\begin{figure*}
    \centering
    \includegraphics[width=0.9\linewidth,page=2,trim={0 8.0cm 0 2.2cm},clip]{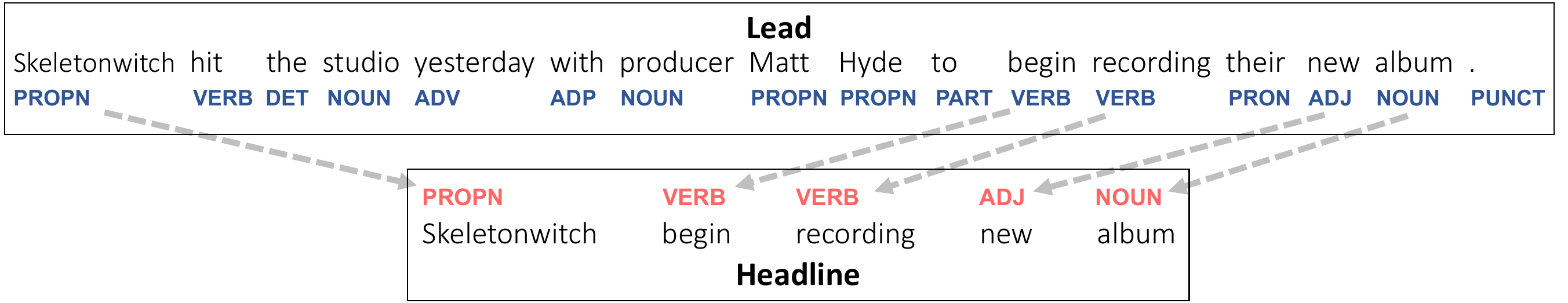}
    \caption{Example of tag projection: tags of lead words are projected onto the corresponding headline word.}
    \label{fig:projection_cartoon}
\end{figure*}

We construct a silver-labeled training set for GSC headlines by selecting articles from the GSC corpus where the headline is a (possibly non-continguous) subsequence of the lead sentence after lowercasing.  After excluding GSCh examples, we find a total of 44,591 out of 200,000 headlines that satisfy this condition, which we divide into 70\% for the training fold and 30\% for a validation fold.  We run inference with the selected EWT-trained tagger on these lead sentences, and assign the predicted tag for each word in the lead sentence to the corresponding aligned word in the headline.  These silver-labeled headlines are used both for training and model selection (GSCproj). See \Cref{fig:projection_cartoon} for an example of this projection. 

This approach was inspired by work on projecting syntactic and morphological annotations across languages \cite{yarowsky-etal-2001-inducing, fossum-abney-2005-automatically,tackstrom2013token,buys2016cross,eskander2020unsupervised}, except that we impose no type constraints on the projected tags in our alignment.  Although \newcite{filippova2013overcoming} state that there are relatively few headlines which are extractive summaries of the article's lead sentence, there were enough of such headlines in the GSC corpus for us to construct a silver-labeled training set of the same order of magnitude as the EWT (12,543 sentences vs. 31,213 headlines).  

\section{Experiments}
\label{sec:experiments}

Models are trained on either EWT, GSCproj, or both training sets. We evaluate all models on the GSCh test set according to token and headline accuracy, unless otherwise stated.
We also consider training multi-domain models on the three auxiliary datasets as well as EWT and GSCproj, to explore the value added by training on more long-form text data.  For the EWT-trained baseline taggers, we insert a final period to GSCh headlines before running inference so as to mitigate the mismatch between the training and test data.

\subsection{Training Details}
\label{subsec:training_details}

We train all models using the Adam optimizer with $\beta_1 = 0.99, \beta_2 = 0.999$.  All BiGRU layers are 100-dimensional.  For each architecture, we perform a random search with a budget of 10 runs over dropout rate in $[0.0, 0.4]$ and number of epochs in $[2, 6]$.  The fixed dropout rate is applied to the hidden layers of the BiGRU, throughout the BERT encoder, and before the tag projection layer.  We sample the exponent of the base learning rate (base 10) for non-contextual taggers uniformly at random from $[-5.5, -1.0]$ and for contextual taggers from $[-5.5, -4.0]$.  Each model was trained using a single V100 GPU and training time varied from one to eight hours based on the size of the training set and number of training epochs.

We retrain each model three times with different random seeds, and select training parameters based on mean GSCproj validation token accuracy across seeds (for models trained solely on EWT, models are selected based on the EWT validation set).  When training jointly on only the EWT and auxiliary datasets, we select a model over hyperparameters and decoder head based on token accuracy on the GSCproj validation set.  We report test performance for the model found to be most accurate on the validation set.  As is recommended by \newcite{sogaard2014s}, we use bootstrap sampling tests to test for statistical significance at the $p < 0.01$ level.\footnote{Using batches of 1056 headlines (20\% of the data) sampled with replacement from GSCh and 2,000 replications.}

\subsection{Extrinsic Open IE Evaluation}
\label{subsec:openie_eval}

To demonstrate the potential impact of explicit modeling of headlines, we consider the effect of POS tag quality on the downstream performance of an Open IE system.
We experiment with a re-implementation of the state-of-the-art Open IE system PredPatt \cite{white2016universal}, which extracts propositions, tuples of predicate followed by $n$ extracted arguments, using syntactic rules
over a UD dependency parse of the source sentence.
For the parser, we employ an arc-hybrid transition-based dependency parser using a bi-directional LSTM neural network ~\cite{kiperwasser-goldberg-2016-simple}, trained on data from sections of 2-21 of the WSJ Penn Treebank.\footnote{The parser achieves 94.47 \% UAS and 93.50 \% LAS on section 23 of WSJ, when using gold POS tags.} The parser takes POS tags as input. In our experiments, we substitute the predicted POS tags from the models under consideration and evaluate their downstream impact on tuple extraction quality.

\removed{

To evaluate this system, we annotated 300 sentences, randomly sampled from GSCh, with tuple extractions, following the general annotation principles from \cite{bhardwaj-etal-2019-carb}, with annotation guideline refinements to address corner cases and headline-specific deviations.  

We use the CaRB evaluation suite from \cite{bhardwaj-etal-2019-carb} to evaluate the Open IE system operating over different POS tags. 
}

To evaluate and compare the systems, we perform a typological error analysis for a randomly sampled 100 sentence subset of GSCh where the output of the OpenIE systems on EWT and EWT+GSCproj-based predictions differ.


\section{Results and Discussion}
\label{sec:discussion}


\begin{table*}
    \centering
    \begin{tabular}{p{3.0cm}|p{3.8cm}|p{1.55cm}|p{1.55cm}|p{1.55cm}|p{1.55cm}}
        \hline & & \multicolumn{2}{c}{\textbf{Non-contextual}} & \multicolumn{2}{c}{\textbf{BERT}} \\ 
        {\bf Setting} & {\bf Train set} & {\bf Token} & {\bf Headline} & {\bf Token} & {\bf Headline} \\ \hline
        & EWT & 90.16 & 53.22 & 92.08 & 60.77 \\
        projected from lead & EWT & 91.71$^{\star}$ & 59.03$^{\star}$ & \textbf{94.00}$^\star$ & \textbf{68.39}$^{\star}$ \\
        & GSCproj & 92.14$^{\star}$ & 60.37$^{\star}$ & 93.56$^{\star}$ & 66.46$^{\star}$ \\
        shared & EWT + GSCproj & 91.94$^{\star}$ & 59.38$^{\star}$ & 93.86$^{\star}$ & 67.76$^{\star}$ \\
        multi-domain & EWT + GSCproj & 92.26$^{\star}$ & 60.34$^{\star}$ & 93.93$^{\star}$ & 68.27$^{\star}$ \\
        multi-domain & EWT + Aux & 92.02* & 59.03* & 92.69* & 63.13 \\
        multi-domain & EWT + GSCproj + Aux & \textbf{93.19}$^{\star\dagger}$ & \textbf{64.01}$^{\star\dagger}$ & \textbf{94.00}$^{\star}$ & 68.10$^{\star}$ \\ \hline
    \end{tabular}
    \caption{Token and headline percent accuracy of each POS tagger on GSCh.  \emph{Aux} refers to all auxiliary datasets described in \Cref{subsec:aux_data}.  \emph{projected from lead} refers to performance attained by predicted tags projected from the lead sentence to headline.  \emph{shared} and \emph{multi-domain} refer to whether the decoder is shared across registers, or decoder weights are specific to each register.  The best performance for each type of encoder is in bold. Statistically significant performance over the EWT and GSCproj models are denoted by $^\star$ and $^\dagger$, respectively.}
    \label{tab:model_performance}
\end{table*}

\Cref{tab:model_performance} displays the test performance for POS taggers on GSCh.  The selected hyperparameters and validation performance of each model is given in Appendix \ref{app:val_perf}. Both non-contextual and contextual language models benefit from training on the silver-labeled GSCproj.  Although the BERT EWT tagger achieves 95.40\% accuracy on the EWT test set, which is on par with strong models such as Stanza \cite{qi2018universal}, the performance degrades on GSCh, achieving only 92.08\% accuracy.  We found that this is in line with performance of an EWT-trained tagger applied to data from other domains: 94.34\% token accuracy on the UD-converted English news treebank test set vs. 97.64\% for a POS tagger trained on the Revised English News Text Treebank.
This underscores the fact that the language and POS tag distribution in news headlines is substantially different from long-form text.  However, by training on GSCproj alone, one can improve absolute accuracy by 1.48\% at the token level and 5.69\% at the headline level.

We find that projecting predicted tags from the BERT tagger from the lead sentence onto the headline performs surprisingly well.  However, this is not a realistic scenario at inference time as few articles have headlines that are subsequences of the lead sentence. We find that a multi-domain BERT tagger trained on five corpora achieves the same performance on GSCh as the POS tags projected from the lead sentence.\footnote{Note that both the EWT+GSCproj and EWT+GSCProj+Aux multi-domain BERT taggers also achieve statistically significantly better token accuracy than the GSCproj model, but only at the $p<0.05$ level.}

\paragraph{Non-contextual performance} Non-contextual taggers also benefit from training on GSCproj.  In fact, a multi-domain non-contextual tagger that is trained on EWT and GSCproj along with the auxiliary datasets outperforms a BERT tagger that was only trained on EWT by over 1\% absolute token accuracy.  Although adding auxiliary long-form text datasets improves tagger accuracy over only training on EWT, the addition of the silver-labeled GSCproj training set leads to more than a 1\% absolute improvement in token accuracy.  One interesting difference between contextual and non-contextual taggers is that the domain-agnostic (\emph{shared}) BERT tagger benefits from the concatenation of EWT and GSCproj training sets, while the non-contextual tagger does not.  We posit that this is because the contextual token embeddings learned by the BERT model naturally discriminate between the headline and long-form text registers.  In effect, the BERT tagger can learn a register-specific sense of each word simply due to distributional differences between the GSCh and EWT registers.

\paragraph{Performance on general headlines} \Cref{tab:nytgsc_evalset_perf} shows the performance of trained BERT models on the NYT+GSC unconstrained headline evaluation set described in \Cref{subsec:aux_data}.  Training on GSCproj alone improves the performance on the GSCh subset, but the model achieves similar token accuracy on the NYT subset.  However, the multi-domain model trained on all available domains yields an absolute increase of 1.67\% token accuracy on the NYT fold, even though none of the training domains explicitly contain NYT headlines.
This suggests that training on projected tags also improves headline POS tagging for headlines that are not strictly subsequences of the lead, or are drawn from a different wire, such as the New York Times.

\begin{table}[t]
    \centering
    \begin{tabular}{c|c|l|l}
        \hline
         &  & \multicolumn{2}{c}{\textbf{\% Accuracy}} \\
         {\bf Train} & {\bf Test} & {\bf Token} &  {\bf Head} \\
        \hline
        \multirow{3}{*}{EWT} & NYT+GSCh & 91.88 & 58.20 \\
         & NYT & 92.14 & 60.26 \\
         & GSCh & 91.59 & 56.46 \\ \hline
        \multirow{3}{*}{GSCProj} & NYT+GSCh & 92.82 & 61.20 \\
         & NYT & 92.03 & 56.77 \\
         & GSCh & 93.59 & 65.68 \\ \hline
        \multirow{3}{*}{\parbox{2cm}{\centering EWT+ GSCProj+ Aux}} & NYT+GSCh & 93.64 & 65.20 \\
         & NYT & 93.81 & 66.38 \\
         & GSCh & 93.43 & 63.84 \\ \hline
    \end{tabular}
    \caption{Performance of models on the general headline evaluation set.  The "EWT+GSCProj+Aux" model is the multi-domain model, and evaluation set predictions are made using the GSCProj head.  Performance is reported separately for the NYT and GSCh subsets, as well as their union.}
    \label{tab:nytgsc_evalset_perf}
\end{table}

\subsection{Training with Gold Supervision}
\label{subsec:train_on_gold}

In the above experiments, we assume no gold in-domain supervision, only training on projected tags.  To better understand how well a model performs with gold supervision, we also train a single-domain contextual tagger using the gold GSCh tags using 5-fold cross validation: training on 60\%, validating on 20\%, and testing on the remaining 20\%.  This tagger, trained on 3,149 gold headlines, achieves 93.55\% token and 66.83\% headline CV test accuracy, which is similar to that achieved by training on the 31,213 GSCproj training set (93.56\%).  A model CV-trained on 3,149 projected tag headlines instead achieves 92.73\% token and 62.73\% headline accuracy.

\subsection{Error Analysis}
\label{subsec:error_analysis}

\begin{figure*}[t]
    \centering
    \hspace{0.6 cm}
    \includegraphics[width=0.36\linewidth,page=6,trim={1.6cm 1.2cm 1.6cm 8.5cm},clip]{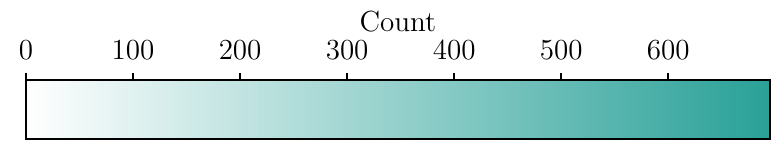} \hspace{1.65 cm}
    \includegraphics[width=0.36\linewidth,page=14, trim={1.6cm 1.2cm 1.6cm 8.5cm},clip]{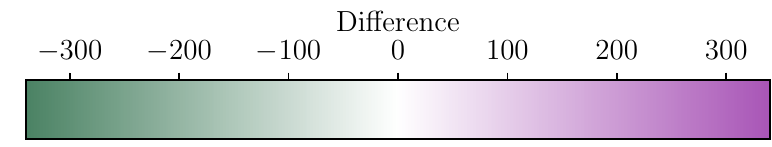}
    \includegraphics[width=0.485\linewidth,page=5,trim={1.6cm 0.2cm 3cm 0.2cm},clip,valign=c]{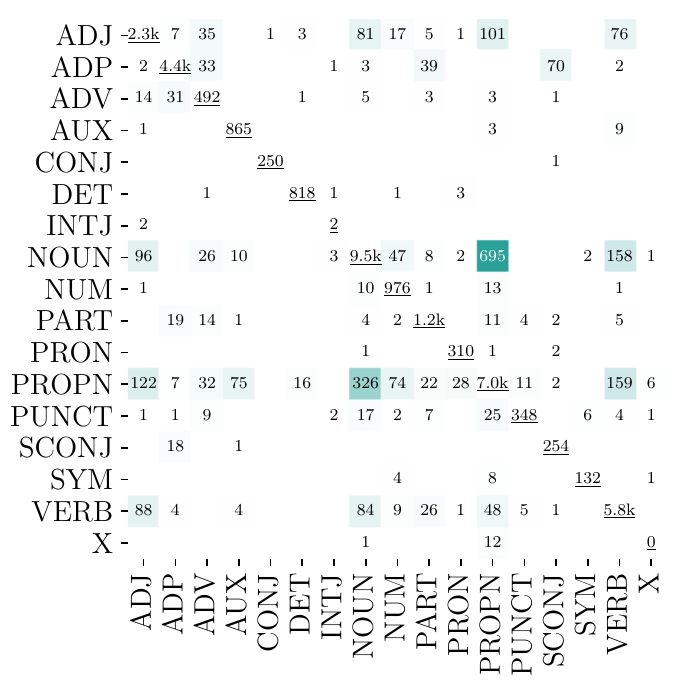}
    \includegraphics[width=0.485\linewidth,page=13,trim={1.6cm 0.2cm 3cm 0.2cm},clip,valign=c]{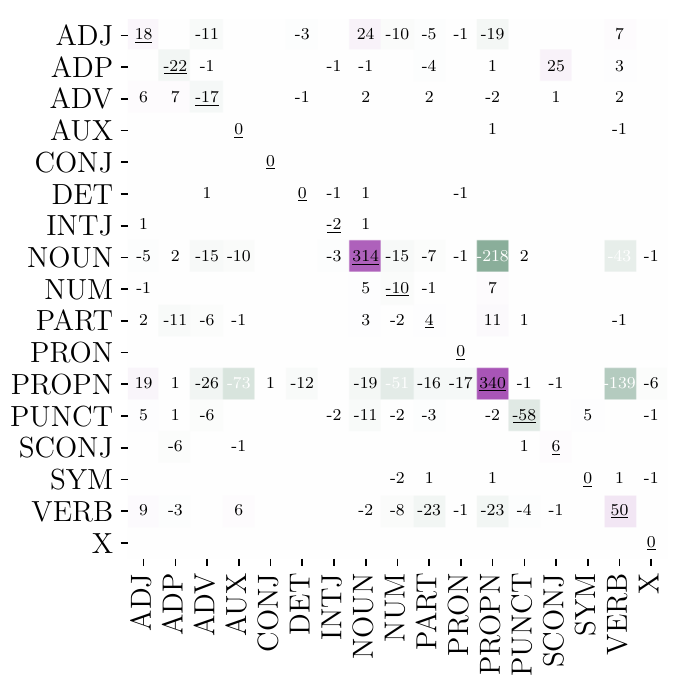}
    \caption{Confusion matrix for the EWT baseline (left) and the difference between the EWT+GSCproj and EWT tagger confusion matrices (right).  In the right plot, purple cells correspond to cases where the EWT+GSCproj model made a prediction more frequently than the EWT model, and vice-versa for green cells.  Zero count cells are not annotated and the diagonal counts (correct predictions) are underlined.}
    \label{fig:ewt_gsch_confusion}
\end{figure*}

The multi-domain EWT+GSCproj tagger improves over the EWT baseline by remedying errors related to the idiosyncracies of headlines. One such error is the frequent confusion of \noun{} and \propn{} tags by the baseline model (\Cref{fig:ewt_gsch_confusion}). This is to be expected as determiners are frequently omitted in headlines, removing a clear signal of whether a word is a \propn.  Take this prediction from the baseline tagger:

\exg. Firefighter accused of setting fires\\
*\propn{} \posverb{} \adp{} \posverb{} \noun\\

The second set of remedied errors are a product of the way that some verb phrases are formed in headlines, omitting the copula and using the \texttt{to \posverb{}} construction to signal  future tense.  These phenomena cause the baseline tagger to make particularly egregious blunders, such as tagging proper nouns as \aux{} when preceding a non-finite \posverb:

\exg.  Hideo Kojima working on a new game\\
\propn{} *\aux{} \posverb{} \adp{} \posdet{} \adj{} \noun\\

or \posverb{} if it precedes \texttt{to \posverb}:

\exg.  Centrelink to review procedures\\
*\posverb{} \aux{} \posverb{} \noun\\


We initially posited that this was due to the prior over label sequences learned by the CRF layer for the EWT register trumping the evidence.  However, when we retrain the EWT model without a CRF layer it only achieved 91.90\% accuracy on GSCh, performing worse than including a CRF layer.

 The multi-domain tagger also remedies a smaller set of errors involving lexical ambiguity; for instance, confusing a nominal for a \posverb:

\exg. Bosc faces spell on sidelines\\
\propn{} \posverb{} *\posverb{} \adp{} \noun\\

Although the multi-domain model occasionally confuses \adp{} for \sconj, after reviewing these examples, we find that many of the errors are due to noisy annotations where a word was incorrectly annotated as \adp{} instead of \sconj.  We also notice annotation inconsistency in constructions such as \texttt{\textbf{(set|engaged|surprised|thrilled)} to} where the gold label should be \adj, not \posverb.

\NoteAB{May want to include this in footnote if saving space.}

\subsection{Extrinsic Evaluation}
\label{subsec:results_openie_eval}

\begin{figure}[h]
    \centering
    \includegraphics[width=\linewidth,page=6,trim={2.5cm 1cm 0.5cm 1cm},clip]{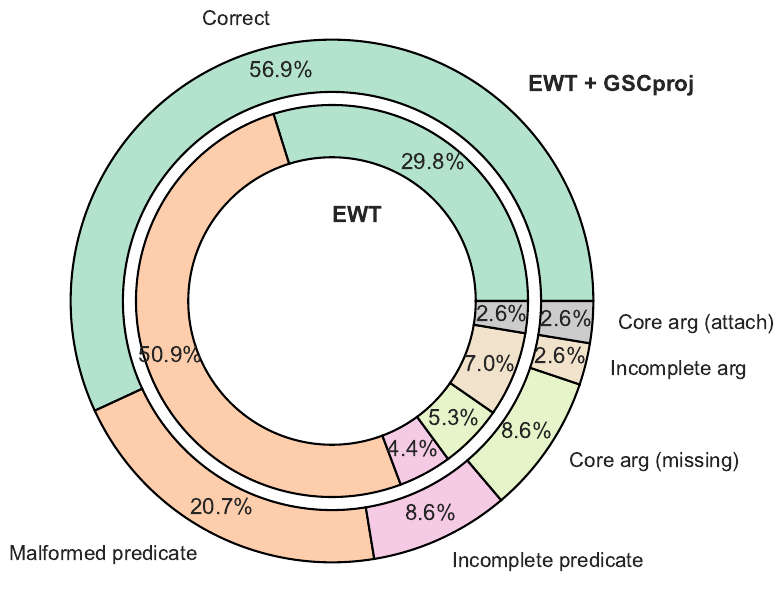}
    \caption{Error types in OpenIE tuple extractions given the POS tags predicted by the EWT (inner donut) and EWT+GSCproj (outer donut) BERT taggers.}
    \label{fig:openie_error_analysis}
\end{figure}


In the extrinsic Open IE-based evaluation, we performed an error analysis of tuples extracted by PredPatt using BERT EWT vs. EWT+GSCproj tags from 100 randomly selected headlines (which contained at least one tuple extracted from each and for which the extractions differed).  Extracted tuples are labeled by two annotators who achieved an inter-annotator agreement rate of 90.0\% before remedying discrepancies. 
Each extracted tuple is labeled for validity, and if invalid, the type of the salient error. 114 and 116 tuples are extracted by the baseline and multi-domain models, respectively. The types of salient error we focus on are:  1. an argument is mis-attached to the predicate; 2. the argument extraction is incomplete; 3. a core argument is missing; 4. the predicate is malformed; or 5. the predicate is incomplete (\Cref{tab:jazzy_openie_error_types}). Our typology is informed by previous work on Open IE evaluations~\cite{schneider-etal-2017-analysing,bhardwaj-etal-2019-carb}. 


\begin{table*}
    \centering
\begin{center}
\begin{tabular}{ p{4.1cm}|p{10cm} } 
 \hline
 \textbf{Error Type} & \textbf{Example Extraction} \\
 \hline
 \multirow{1}{*}{\textbf{Argument Misattachment}} & 
 {\small Joliet man accused of shooting \argone{55 dogs} \predicate{released from} \argtwo{hospital}} \\ 
 \multirow{1}{*}{\textbf{Incomplete Argument}} &
{\small \argone{Schoolground knife} and weapon attacks have \predicate{doubled in} \argtwo{10 years}} \\ 
\multirow{1}{*}{\textbf{Missing Core Argument}} &
{\small \argone{Gene therapy} \predicate{boosts} hope for the deaf} \\ 
\multirow{2}{*}{\textbf{Malformed Predicate}} &
{\small \argone{Newcomers} will need to learn \predicate{about military}} \\
& {\small \predicate{Affymetrix to acquire} \argone{Panomics}} \\ 
\multirow{1}{*}{\textbf{Bad Sub-predicate}} & 
{\small \argone{Maoists} \predicate{set} \argtwo{liquor unit} afire} \\
\hline
\end{tabular}
\end{center}
\caption{OpenIE error types.  Extracted tuples encoded as $\langle$\predicate{Predicate}, \argone{Argument 1}, \argtwo{Argument 2}$\rangle$.}
    \label{tab:jazzy_openie_error_types}
\end{table*}

There are two striking differences between tuples extracted using the baseline EWT and EWT + GSCproj POS tags.  First, the multi-domain model results in far more precise extractions than the baseline, achieving a precision of 56.9\% vs. 29.8\%.  Second, the majority of errors resulting from the baseline tagger are due to malformed predicates.  Unlike error types such as ``incomplete argument'' or ``core argument (attachment)'', these errors are particularly egregious, as the essential meaning of the extracted tuple is corrupted.  Many instances of malformed predicates are due to part of the subject being improperly placed in the predicate, a byproduct of the baseline tagger's tendency to identify words in second position or preceding ``to'' as \aux{} or \posverb.

\NoteAB{Need description of error typology in Experiment section.}

\NoteAB{Insert examples of malformed predicate, how this relates to AUX/VERB misclassifications by EWT in error analysis above.}

\section{Related Work}
\label{sec:related_work}

``Headlinese'' has been identified as a unique register at least since \newcite{straumann1935newspaper}.  
Hallmarks of English headlinese include the omission of articles and auxiliary verbs and using the infinitival form of verbs for future events.  Subsequent work has found that, even within the English language, the syntax of news headlines varies as a function of publication \cite{maardh1980headlinese,siegal1999new}, time period \cite{vanderbergen1981grammar,schneider2000emergence,afful2014diachronic}, and country \cite{ehineni2014syntactic}. In this work, we primarily consider headlines from the GSC, drawn from over ten thousand English language news sites in 2012, and leave investigating syntactic variation as future work.

Most NLP work on news headlines has focused on the problem of headline generation \cite{banko2000headline,rush2015neural,takase2016neural,tan2017neural,takase2019positional} or summarization \cite{filippova2013overcoming}.
There has also been work on training headline classification models to label headlines by their expressed emotion \cite{kozareva2007ua,oberlander2020goodnewseveryone}, stance with respect to an issue \cite{ferreira2016emergent}, framing of/bias towards a political issue \cite{gangula2019detecting,liu2019detecting}, or the category or value of the news article \cite{di2017predicting}.  However, we are not aware of work to develop headline-specific models for predicting traditional linguistic annotations such as POS tags or dependency trees, although there has been work on adapting machine translation models to headlines \cite{ono2003translation}. As we show in this work, and has been observed in the past \cite{filippova2013overcoming}, a model trained on non-headline-domain data will not necessarily perform well on headlines.

Our work is analogous to work on constructing POS taggers or syntactic parsers for less traditional domains such as tweets \cite{owoputi2013improved,kong2014dependency,liu2018parsing}.  Unlike this line of work, we did not have to craft a unique tag set for headlines, as headlines are not rife with the typographical errors and atypical constructions of tweets.  At the same time, we do not presume a gold-annotated headline training set, but use projection to construct a silver-labeled training set.  Data is annotated data purely for the purpose of evaluation.

\section{Conclusion}
\label{sec:conclusion}

This work is a first step towards developing stronger NLP tools for news headlines.  We show that training a tagger on headlines with projected POS tags results in a far stronger model than taggers trained on gold-annotated long-form text. This suggests that more expensive syntactic annotations, such as dependency trees, may also be reliably projected onto headlines, obviating the need for gold dependency annotations when training a headline parser.

Although this work is focused on learning strong headline POS taggers, the projection technique described here can be adapted to train other strong headline sequence taggers; for example, training a headline chunker or named entity tagger on IOB tags projected from the lead sentence.
Projection could potentially be applied to generate silver-labeled data for other domains such as simplified English (e.g., aligned sentences from simplified to original Wikipedia \cite{coster2011simple}) and other languages.


\section{Acknowledgments}
\label{sec:acknowledgments}

Many members of the Bloomberg AI group gave helpful feedback on early drafts of this work. In particular, this work benefited immensely from discussions with Amanda Stent, as well as aesthetic feedback and figure-tinkering from Ozan \.{I}rsoy.

\bibliographystyle{acl_natbib}
\bibliography{main}

\clearpage
\appendices

\section{Annotation Guidelines}
\label{app:annotation_guidelines}

\begin{figure*}
    \centering
    \includegraphics[width=0.52\linewidth]{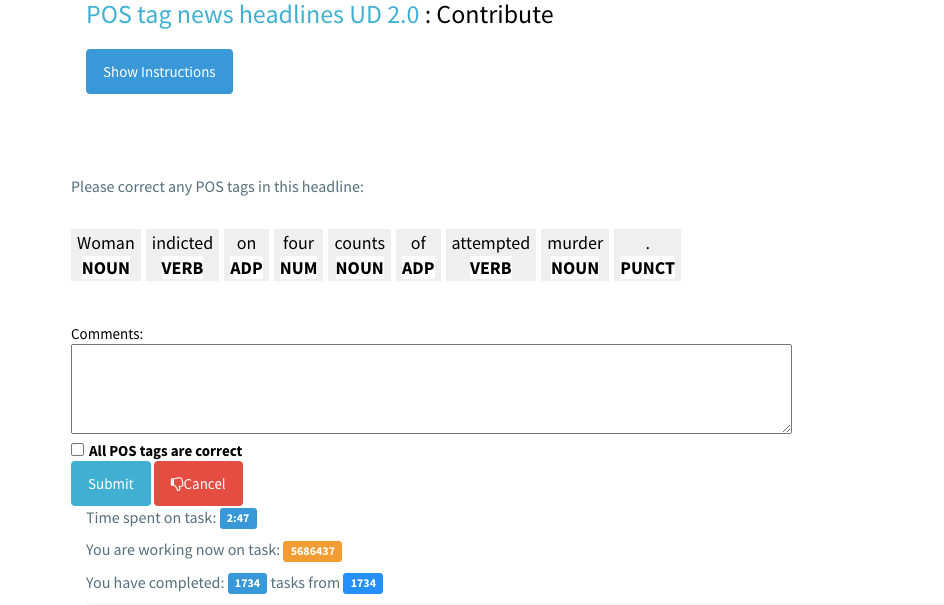}
    \includegraphics[width=0.44\linewidth]{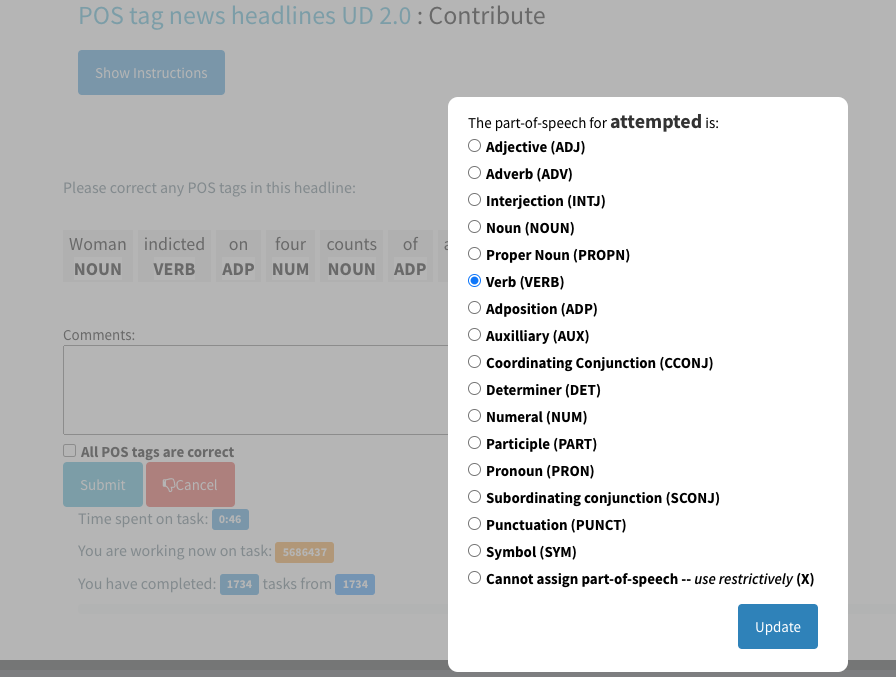}
    \caption{Example of the annotation UI initially (left) and after selecting a POS tag to correct (right).  The instructions are hidden.}
    \label{fig:crowdsourcing_ui}
\end{figure*}

The annotation UI is shown in \Cref{fig:crowdsourcing_ui}.  See below for full instructions for the annotation task.  When in doubt, annotators consulted the UD POS tagging guidelines at \url{https://universaldependencies.org/u/pos/} and similar sentences in the EWT.

\subsection*{Annotation Guidelines for POS Tagging Headlines}

For each task, you will be shown a headline along with POS tags from the Universal Dependencies (UD) 2.0 schema. These tags are automatically assigned by a trained model.

Each word in the headline has its assigned POS tag displayed underneath. Click on any word to correct its tag. POS tags that you changed will be marked with a red border. If all of the words are assigned the correct tag, be sure to select the "All POS tags are correct" checkbox, otherwise you will not be able to click Submit and proceed to the next task.

You should assign POS tags according to the directions given in the UD 2.0 guidelines. Be sure to read these guidelines thoroughly before beginning annotation. When in doubt, you can search for similar sentences in the English Web Treebank here: \url{http://bionlp-www.utu.fi/dep_search/} (select "English (UDv2.0)" and note their parts of speech).

\subsubsection*{Tricky Cases}

There are a few classes of mistakes our model makes consistently. There are also a few classes of examples that may be difficult for humans to decide which is the correct POS tag. Below are the most common cases we encountered.

\paragraph*{NOUN vs. PROPN}

Our model will often incorrectly label \noun{} tokens as \propn, because headlines often do not precede nouns with a determiner. These are easy errors for humans to fix, but just be aware that they are common. Example:

\ex. Haye hopes fight is only postponed

"Haye" should be tagged as a \propn{} in this headline, not \noun.

\paragraph*{Compound Nouns}

From the UD 2.0 guidelines: "A noun modifying another noun to form a compound noun is given the tag \noun{} not \adj." Example:

\ex. But air traffic controllers in Baghdad have no record of the flights, which supposedly took off between July 2004 and July 2005.

In this sentence, "air", "traffic", and "controllers" should all be labeled as \noun. When in doubt, defer to what the typical dictionary POS tag for a word would be.

\paragraph*{Multiword Proper Nouns}

All tokens constituting a multiword name that overall functions like a proper noun should be tagged as \propn, even if each constituent word would typically be given another POS tag. Examples:

\exg. United Airlines\\
\propn{} \propn{}\\

\exg. The Yellow Pages\\
\propn{} \propn{} \propn\\

These are typically names of organizations or people. This also goes for titles preceding a \propn. Examples:

\exg. US Rep. Lamborn\\
\propn{} \propn{} \propn\\

\exg. State Senator Stewart Greenleaf\\
\propn{} \propn{} \propn{} \propn\\

\exg. Former President Bush\\
\adj{} \propn{} \propn\\

In the above, "Former" is tagged as an \adj, since it is not considered part of the title. One major exception is the possessive marker, "'s", which is tagged as \pospart, a participle, even if it is part of an organization's name. Examples:

\exg. Pakistan People 's Party\\
\propn{} \propn{} \pospart{} \propn\\

\exg. Garden Grove seniors invited to  Mother 's Day event\\ \propn{} \propn{} \noun{} \posverb{} \adp{} \propn{} \pospart{} \propn{} \noun\\

Some corner cases:

Tokens that consist of all digits should be labeled as \num{} (e.g., the 7 in Windows 7).
Acronyms of proper nouns should be labeled as \propn: USA, NATO, HBO.
When in doubt, refer to similar sentences in the English Web Treebank.

\paragraph*{Currencies}

Tagging currencies can be tricky. If the currency denotes the denomination of a sum of money, it should be marked as a \sym:

\exg. Rs. 500 in  Pakistan\\
\sym{} \num{} \adp{} \propn\\

\exg. \$   500 in  USA\\
\sym{} \num{} \adp{} \propn\\

\exg. kr  500 in  Denmark\\
\sym{} \num{} \adp{} \propn\\

A currency that is referred to generically should be labeled as a \noun:

\exg. Yuan was little changed against the dollar\\
\noun{} \aux{} \adv{} \posverb{} \adp{} \posdet{} \noun\\

\exg. these countries could put  the dollar under intense pressure\\ \posdet{} \noun{} \aux{} \posverb{} \posdet{} \noun{} \adp{} \adj{} \noun\\

\exg. the Japanese government holds dollar reserves of  approximately \$   1   trillion\\
\posdet{} \adj{} \noun{} \posverb{} \noun{} \noun{} \adp{} \adj{} \sym{} \num{} \num\\

\paragraph*{Hyphenated Compounds and Multiword Expressions}

Multiword expressions are phrases made up of at least two words whose meaning cannot be directly inferred from of its constituent words. A canonical example is the phrase "kick the bucket", which means "to die", and can take the place of a verb in a sentence. Each word in a multiword expression should be assigned the POS tag that is typical for its usage. Example:

\exg. let  the cat  out of  the bag\\
\posverb{} \posdet{} \noun{} \adp{} \adp{} \posdet{} \noun\\

Similarly, hyphenated expressions like "drive-through" are split by punctation into separate tokens ("drive", "-", "through"). POS tags should be assigned to each constituent token independently. Example:

\exg. drive - through\\
\posverb{} \punct{} \adp\\

When in doubt, defer to the typical POS tag for a token rather than the context.

\paragraph*{Passive Constructions}

Be aware of passive constructions when the subject of the verb is omitted.

\ex. Tourism Industry is least affected

\ex. Grand - standing is ignored .

In these cases, "affected" and "ignored" should both be labeled as \posverb, although one might be tempted to label them as \adj.

\paragraph*{Copula is AUX}

Copulas (is/are/some inflection of to be) are always labeled as \aux, even when used as the main verb in a sentence. Example:

\ex. What are catch shares ?

"are" should be tagged as \aux{} here, not \posverb.

\paragraph*{Tagging "than"/"for" SCONJ vs. ADP}

Words like "than" and "for" may be particularly difficult to tag. When these words are followed by a noun phrase, they are likely \adp. When they are followed by a clause, they are likely \sconj. Examples:

\exg. Meiring is  more than an  embarrassment . \\
\propn{} \aux{} \adj{} \adp{} \posdet{} \noun{} \punct\\

\exg. The Kashmir issue generates far more terrorism ,     and   even the threat of  nuclear war  ,     than  Iraq  ever did  .\\
\posdet{} \propn{} \noun{} \posverb{} \adv{} \adj{} \noun{} \punct{} \cconj{} \adv{} \posdet{} \noun{} \adp{} \adj{} \noun{} \punct{} \sconj{} \propn{} \adv{} \posverb{} \punct\\

\exg. Now the Electoral Commission is  refusing to   punish people for mere past Baath Party membership .\\
\adv{} \posdet{} \propn{} \propn{} \aux{} \posverb{} \pospart{} \posverb {} \noun{} \adp{} \adj{} \adj{} \propn{} \propn{} \noun{} \punct\\

\section{Model Validation Performance}
\label{app:val_perf}

\begin{table*}[t]
    \centering
    \begin{tabular}{c|c|l|l|l|l}
        \hline
        {\bf Encoder Type} & {\bf Model} & {\bf LR} & {\bf DR} & {\bf Epochs} & {\bf \% Token Acc} \\
        \hline
        \multirow{6}{*}{Non-contextual} & EWT & 3.49e-03 & 0.0438 & 2 & 93.52 ($\pm 0.0549$) \\
         & GSCproj & 3.12e-04 & 0.265 & 5 & 93.70 ($\pm 0.0618$)\\
         & EWT+GSCproj[shared] & 1.29e-03 & 0.315 & 5 & 93.83 ($\pm 0.0462$)\\
         & EWT+GSCproj[multi] & 2.95e-04 & 0.0763 & 5 & 93.38 ($\pm 0.128$)\\
         & EWT+Aux[multi] & 6.43e-05 & 0.118 & 6 & 90.44 ($\pm 0.120$)\\
         & EWT+GSCproj+Aux[multi] & 8.71e-05 & 0.327 & 6 & 93.78 ($\pm 0.0224$)\\ \hline
        \multirow{6}{*}{BERT} & EWT & 1.77e-05 & 0.262 & 6 & 95.58 ($\pm 0.0519$) \\
         & GSCproj & 5.22e-06 & 0.0519 & 6 & 95.66 ($\pm 0.0523$) \\
         & EWT+GSCproj[shared] & 4.27e-06 & 0.191 & 5 & 95.91 ($\pm 0.0403$) \\
         & EWT+GSCproj[multi] & 6.71e-06 & 0.276 & 5 & 95.93 ($\pm 0.0494$)\\
         & EWT+Aux[multi] & 3.57e-06 & 0.244 & 6 & 93.35 ($\pm 0.0303$) \\
         & EWT+GSCproj+Aux[multi] & 1.73e-05 & 0.164 & 4 & 95.72 ($\pm 0.0616$)\\ \hline
    \end{tabular}
    \caption{Hyperparameters (\emph{LR}: base learning rate, \emph{DR}: dropout rate, and number of training \emph{Epochs}) and validation mean \% token accuracy $(\pm $ standard deviation) for models presented in \Cref{tab:model_performance}.  Standard deviation and mean token accuracy are computed over three training runs with different random seeds.}
    \label{tab:selected_params}
\end{table*}

Here we report the selected hyperparameters and validation performance for models presented in \Cref{tab:model_performance}.  Note that validation performance is over the silver-labeled GSCproj dataset, which means that it is likely an overestimate of the performance on the gold validation set.  The one exception to this is the EWT baseline tagger, where validation performance is over the EWT validation set.  \Cref{tab:selected_params} gives the selected hyperparameters for each of these models along with their validation performance.  For the EWT+Aux tagger, validation performance is reported for the EWT decoder for both non-contextual and BERT taggers, as this decoder achieved the highest token accuracy on GSCproj.

\REMOVED{
\section{Performance on General Headlines}
\label{app:nytgsc_evalset_perf}

\Cref{tab:nytgsc_evalset_perf} shows the performance of trained BERT models on the NYT+GSC unconstrained headline evaluation set.  Training on GSCproj alone improves the performance on the GSCh subset, but the model achieves similar token accuracy on the NYT subset.  However, the multi-domain model trained on all available domains yields an absolute increase of 1.67\% token accuracy on the NYT fold, even though none of the training domains explicitly contain NYT headlines.

\begin{table}[t]
    \centering
    \begin{tabular}{c|c|l|l}
        \hline
         &  & \multicolumn{2}{c}{\textbf{\% Accuracy}} \\
         {\bf Train} & {\bf Test} & {\bf Token} &  {\bf Headline} \\
        \hline
        \multirow{3}{*}{EWT} & NYT+GSCh & 91.88 & 58.20 \\
         & NYT & 92.14 & 60.26 \\
         & GSCh & 91.59 & 56.46 \\ \hline
        \multirow{3}{*}{GSCProj} & NYT+GSCh & 92.82 & 61.20 \\
         & NYT & 92.03 & 56.77 \\
         & GSCh & 93.59 & 65.68 \\ \hline
        \multirow{3}{*}{\parbox{2cm}{\centering EWT+ GSCProj+ Aux}} & NYT+GSCh & 93.64 & 65.20 \\
         & NYT & 93.81 & 66.38 \\
         & GSCh & 93.43 & 63.84 \\ \hline
    \end{tabular}
    \caption{Performance of models on the general headline evaluation set.  The "EWT+GSCProj+Aux" model is the multi-domain model, and evaluation set predictions are made using the GSCProj head.  Performance is reported separately for the NYT and GSCh subsets, as well as their union.}
    \label{tab:nytgsc_evalset_perf}
\end{table}
}

\end{document}